\pdfoutput=1

\documentclass[11pt]{article}

\usepackage[preprint]{acl} 

\usepackage{times}
\usepackage{latexsym}
\usepackage{todonotes}
\usepackage{booktabs}
\usepackage{adjustbox}
\usepackage{longtable}
\usepackage[T1]{fontenc}

\usepackage[utf8]{inputenc}

\usepackage{microtype}
\usepackage{multirow}
\usepackage{inconsolata}

\usepackage{graphicx}
\usepackage{enumitem}
\usepackage{colortbl}
\title{Better Alignment with Instruction Back-and-Forth Translation}

\author{Thao Nguyen$^{1,2}$ \ \ \ Jeffrey Li$^1$ \ \ \ Sewoong Oh$^1$ \\ \textbf{Ludwig Schmidt}$^1$ \ \ \ \textbf{Jason Weston}$^2$ \ \ \ \textbf{Luke Zettlemoyer}$^{1,2}$ \ \ \ \textbf{Xian Li}$^2$ \\
$^1$University of Washington \ \ \ $^2$Meta FAIR \\
\small \texttt{\{thaottn,jwl2162,sewoong,schmidt\}@cs.washington.edu} \ \ \ \texttt{\{jase,lsz,xianl\}@meta.com}
}

\begin{document}
\maketitle
\begin{abstract}
We propose a new method, instruction back-and-forth translation, to construct high-quality synthetic data grounded in world knowledge for aligning large language models (LLMs). Given documents from a web corpus, we generate and curate synthetic instructions using the backtranslation approach proposed by \citet{li2023self}, and rewrite the responses to improve their quality further based on the initial documents. Fine-tuning with the resulting (backtranslated instruction, rewritten response) pairs yields higher win rates on AlpacaEval than using other common instruction datasets such as Humpback, ShareGPT, Evol-Instruct, Open Orca, Alpaca-GPT4 and Self-instruct. We also demonstrate that rewriting the responses with an LLM outperforms direct distillation, and the two generated text distributions exhibit significant distinction in embedding space.
Further analysis shows that our backtranslated instructions are of higher quality than other sources of synthetic instructions, while our responses are more diverse and complex than those obtained from distillation. Overall we find that instruction back-and-forth translation combines the best of both worlds---making use of the information diversity and quantity found on the web, while ensuring the quality of the responses which is necessary for effective alignment.
\end{abstract}

\section{Introduction}\label{intro}
\vspace{-0.25em}
\begin{figure*}[t]
\centering
\hspace{-0.1em}
\includegraphics[trim=0 0cm 0 0,clip,width=1.01\linewidth]
{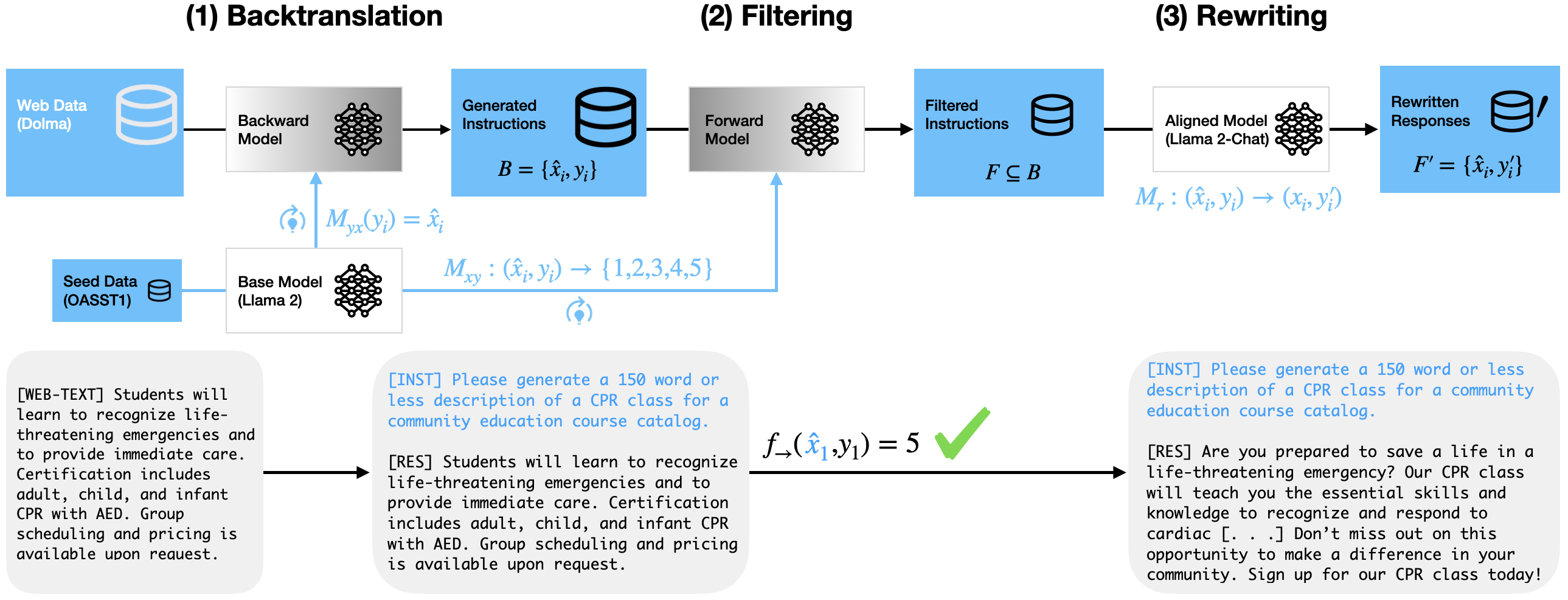}
\vskip -0.5em
\caption{\textbf{Overview of our proposed pipeline: instruction back-and-forth translation.} (1) We first fine-tune a base model, e.g. Llama-2, on some seed examples from Open Assistant, for the task of instruction generation. We then extract initial candidate responses from a web corpus, e.g. Dolma, and use the fine-tuned model to obtain synthetic instructions that would go with the corresponding responses; (2) We separately obtain an instruction-following model by fine-tuning the same base model on the seed examples, and use it to score the quality of the (synthetic instruction, web-scraped response) pairs; (3) With the highest scoring pairs, we ask an existing aligned model (e.g. Llama-2-chat) to improve the responses further, conditioned on the generated instructions and the initial web texts. Steps (1) and (2) follow \citet{li2023self} with some modifications (i.e. using preprocessed documents from Dolma instead of parsing raw HTMLs from ClueWeb). We provide a specific data example in the bottom row.
}
\label{fig:main_fig}
\vspace{-1em}
\end{figure*}
In recent years, it is increasingly common for large language models (LLMs) to be deployed through a chat interface to interact with users' queries. This capability is achieved by taking models that have been pre-trained on massive amounts of web-crawled text and fine-tuning them on a relatively smaller set of instruction-response pairs or preferences \cite{ouyang2022training}. Popular
instruction-tuning corpora are often constructed by \textit{(i)} human annotation and curation \cite{kopf2024openassistant,DatabricksBlog2023DollyV2,zhou2024lima}, \textit{(ii)} converting existing texts, e.g. from other NLP tasks \cite{longpre2023flan} or web crawls \cite{li2023self,koksal2023longform} to instruction-response pairs, and \textit{(iii)} distilling knowledge from a model \cite{vicuna2023,mukherjee2023orca}. 

There are benefits and disadvantages to each of these approaches. While \textit{(i)} can offer strong baselines \cite{zhou2024lima, kopf2024openassistant}, the reliance on human annotations makes scaling up these datasets difficult. 
The success of \textit{(i)} and \textit{(ii)} is also more dependent on having an effective filtering mechanism to remove noisy data from humans and the web. On the other hand, \textit{(iii)} is more cost-effective and scalable; many current state-of-the-art instruction datasets employ distillation. However, distillation alone has also been claimed to be a ``false promise''---\citet{gudibande2023false} shows that specific capability improvements depend on the coverage of the distilled data and that the performance gains from distilling can plateau quickly. These findings suggest that less noisy and more diverse data is crucial for instruction-tuning to close the gap between open and closed LLMs.


Given the knowledge breadth that can be found on the internet, generating instruction-response data with backtranslation has been shown to be a promising alternative to distillation \cite{li2023self}. In particular, the authors parse the ClueWeb corpus \cite{overwijk2022clueweb22} for self-contained text segments, train a model specifically for instruction generation on Open Assistant data \cite{kopf2024openassistant}, and generate instructions to go with the text segments. However, this approach relies on high-quality text data provided by ClueWeb, a paid-access corpus containing only the most popular web pages from search results, as well as a list of manually specified rules for parsing raw HTML files into structured responses (which are then used directly for fine-tuning). 

In this work, we also adopt the instruction backtranslation technique but make it more accessible and scalable. We make direct use of preprocessed documents from a large-scale \textit{open-source} corpus like Dolma \cite{dolma} and generate instructions via backtranslation accordingly. We find that the quality of our instructions are comparable to those backtranslated from ClueWeb. To make up for the lack of manually designed rules for structuring the response, we experiment with using an LLM to rewrite the response based on the generated instruction and the initial web text. This also allows us to avoid directly distilling and overfitting to an LLM's knowledge. An overview of our pipeline, which we call \emph{back-and-forth translation}, can be found in Figure \ref{fig:main_fig}. 


Given the same data quantity, fine-tuning Llama-2-70B on the instruction-response pairs from our data generation pipeline improves the AlpacaEval win rate by 3.6\% compared to using the backtranslation data from previous work \cite{li2023self}, and by at least 3.2\% compared to using other existing distillation datasets such as OpenOrca \cite{OpenOrca}, ShareGPT \cite{vicuna2023}, Evol-Instruct \cite{xu2023wizardlm}, Alpaca-GPT4 \cite{peng2023instruction} and Self-instruct \cite{wang2022self} (Section \ref{performance}). By asking a model to rewrite responses based on initial texts extracted from Dolma, we obtain a distribution of responses that interpolates between the original web text distribution and the distribution of outputs distilled from the same model. Fine-tuning on the rewritten responses in turn outperforms fine-tuning on the distilled responses corresponding to the same instructions (Section \ref{rewrite_vs_distill}). In addition, we offer some insights into how instruction backtranslation and response rewriting affect the quality of instructions and responses respectively, especially in comparison with existing data generation methods (Sections \ref{instruction_quality} and \ref{response_quality}). Overall our results suggest that back-and-forth translation offers an effective way to generate instruction-tuning data enriched with diverse information found on the web, while ensuring the quality of the response annotations by having aligned LLMs in the loop.
\vspace{-0.25em}
\section{Method}\label{method}
Figure \ref{fig:main_fig} shows an overview of our pipeline. Here we describe each step in more detail.
\vspace{-0.25em}
\subsection{Background: instruction backtranslation}
Our work is inspired by the backtranslation method from \citet{li2023self}. In this previous work, the authors fine-tune a base language model on some seed instruction-response pairs ${(x_s, y_s)}$ (e.g. from Open Assistant \cite{kopf2024openassistant}) to obtain a backward model $M_{yx} := p(x|y)$ that learns to generate instructions. The authors then extract candidate responses ${y_i}$ from .warc files of a web corpus, ClueWeb, using carefully constructed HTML-parsing rules, and augment the responses with corresponding instructions output by the backward model. This yields a set of candidate ${(\hat{x}_i, y_i)}$ pairs. 

The authors also separately fine-tune the same base language model on the same seed data ${(x_s, y_s)}$ to obtain a forward model $M_{xy} := p(y|x)$ that can follow instructions. This model is then prompted to score candidate ${(\hat{x}_i, y_i)}$ pairs on a 5-point scale. The final instruction-tuning dataset consists of only score-5 examples. The paper provides ablations to show that this curation step is critical to achieving performance gains, especially with increasing instruction data quantity.

The two steps derived from \citet{li2023self} are denoted as \texttt{(1) Backtranslation} and \texttt{(2) Filtering} in Figure \ref{fig:main_fig}.
\vspace{-0.25em}
\subsection{The rewriting process}
A major limitation of previous work \cite{li2023self} 
is the limited availability of high-quality candidate responses. The authors rely on Clueweb \cite{overwijk2022clueweb22} as the source for unlabelled responses $\{y_i\}$, using only highly linked websites visited by a search engine (e.g. Wikipedia, popular news sites) as this offers quality control over the extracted texts $\{y_i\}$. However, (i) ClueWeb requires paid access, (ii) the text segments were extracted from raw HTML format, which requires specific preprocessing and may still result in segments that are sub-optimal as responses.

Our work removes the data access restrictions and preprocessing steps by using cleaned documents from an open-source corpus, Dolma \cite{dolma}, for the initial web-scraped responses. Since these documents come pre-extracted (with all HTML structures removed) and were written for various purposes, they may contain redundant information and the content presentation could also be improved (e.g. by being split into paragraphs). Consequently, we use an LLM to improve these documents to better resemble responses from AI Assistants; this step is denoted as \texttt{(3) Rewriting} in Figure \ref{fig:main_fig}. Conditioned on initial text $y_i$ from Dolma and the corresponding backtranslated instruction $\hat{x}_i$, we prompt an aligned LLM, Llama-2-70B-chat, to rewrite the response to improve its quality ($y_i'$). The full prompt can be found in Appendix \ref{app:prompt}. 

By default, we apply rewriting to ${(\hat{x}_i, y_i)}$ pairs that have passed the filtering stage. However, we also experiment with skipping the filtering step, i.e. rewrite responses for \textit{any} pair even if the forward model finds some web responses and corresponding generated instructions not properly aligned. Overall we find that step \texttt{(3) Rewriting} is more effective compared to \texttt{(2) Filtering}, though using both offers complementary performance benefits. We will elaborate on this in Section \ref{performance}.

\vspace{-0.25em}
\section{Experiment setup}
\subsection{Training details}
\vspace{-0.25em}
\paragraph{Data.} To source the initial web-crawled responses, we use the Common Crawl subset of Dolma v1 \cite{dolma}. This subset has been preprocessed with quality and content filters, in addition to undergoing deduplication. We additionally filter out documents whose lengths are close to exceeding the context length of Llama-2; this removes about 25\% of the Common Crawl subset. Besides, we only use data from the head split (which consists of documents with the best perplexity scores), as preliminary experiments show that this split offers better candidate responses than the middle split (see Table \ref{tab:cc_sources} in Appendix).

For the seed data used to train the forward and backward models, we follow previous work \cite{li2023self} and use 3200 examples from the Open Assistant dataset \cite{kopf2024openassistant}, chosen from the first turn of each conversation tree. Note that the seed data only consists of English language responses that are considered high-quality, based on their human annotated rank (rank 0).
\vspace{-0.25em}
\paragraph{Model.} We fine-tune a Llama-2-70B base model \cite{touvron2023llama2} on the seed data to obtain the forward and backward models used in steps (1) and (2) of our pipeline. The rewriting step employs Llama-2-70B-chat by default. We also experiment with using a smaller model (Llama-2-7B-chat) as well as the forward model from step (2) for rewriting, but we observe that the output quality is worse (Appendix \ref{app:rewrite_ablations}).
For performance evaluation, we fine-tune both the 7B and 70B scales of the Llama-2 base model on the resulting instruction-response pairs in a supervised manner. Specific hyperparameters can be found in Appendix \ref{app:more_train_details}.
\vspace{-0.25em}
\paragraph{Evaluation.} Given a fine-tuned Llama-2 model, we prompt it to respond to 805 questions from the AlpacaEval benchmark \cite{alpaca_eval} and report the model's win rate over text-davinci-003 as evaluated by GPT-4 model. We also adopt the length-controlled win rate evaluation from AlpacaEval 2.0 \cite{dubois2024length}, see Appendix \ref{app:alpaca2} for more details. Performance on other NLP tasks can be found in Appendix \ref{app:other_nlp_evals}.

\subsection{Baselines}\label{baselines}
The other fine-tuning data sources we compare to include:
\begin{itemize}[topsep=0pt, itemsep=0pt, leftmargin=8pt, parsep=2pt]
    \item \textbf{Open Orca} \cite{OpenOrca,mukherjee2023orca}: contains GPT-4-distilled outputs to FLAN tasks \cite{longpre2023flan}, which are converted from existing NLP datasets using manually crafted templates. The tasks have been augmented with prompting to elicit some form of reasoning during distillation.
    \item \textbf{ShareGPT} \cite{vicuna2023}: the data comes from ShareGPT.com, where users shared their own conversation logs with ChatGPT. We only take the first instruction and first response from each conversation for fine-tuning.
    \item \textbf{ClueWeb + filtering} \cite{li2023self}: responses are parsed from HTML files in the ClueWeb corpus \cite{overwijk2022clueweb22} and do not undergo rewriting. Instructions are generated with the backtranslation approach. After preprocessing and two rounds of curation, previous work produces 41.8K instruction-response pairs in total.
    \item \textbf{Self-instruct} \cite{wang2022self}: the instructions, inputs and outputs are generated by GPT-3, bootstrapped from a small set of seed tasks.
    \item \textbf{Alpaca-GPT4} \cite{peng2023instruction}: contains GPT-4-distilled responses to instructions from Alpaca dataset \cite{taori2023alpaca}. Alpaca's instruction generation seeks to improve over the Self-instruct framework by using different prompts and a more advanced model (text-davinci-003).
    \item \textbf{Evol-Instruct} \cite{xu2023wizardlm}: starting from the Alpaca instruction set, this work uses a set of evolution prompts to rewrite the instructions to improve their complexity. This evolution process is repeated multiple times, using an eliminator in between to filter out the failed instructions. Responses are then distilled from ChatGPT.
\end{itemize}
We note that except for the first two, the rest of the baselines employ synthetic instructions. Besides, the majority of these datasets (other than ClueWeb) distill responses from different existing LLMs. We use these GPT-distilled datasets for research-only, non-commercial purposes (i.e. to serve as competitive baselines to compare our method against).

\vspace{-0.25em}
\section{Fine-tuning results}\label{performance}
We validate the effectiveness of our data generation method, by examining the AlpacaEval performance of Llama-2 models fine-tuned on our data versus on other datasets, given the same instruction quantity.
\vspace{-0.25em}
\subsection{7B scale}
We fix the number of samples to be 12.8K and compare fine-tuning Llama-2-7B base model on our backtranslation and rewritten data to fine-tuning on other common baselines described in Section \ref{baselines}. Results are shown in Table \ref{tab:7b_results}.
Below we describe our different data variations in more detail:
\begin{itemize}[topsep=0pt, itemsep=0pt, leftmargin=8pt, parsep=2pt]
\item \textit{Dolma + filtering:} Data obtained from steps (1) and (2) of our pipeline (Figure \ref{fig:main_fig}). We gather initial responses from the Dolma corpus, generate corresponding instructions with backtranslation, and filter out instruction-response pairs that do not get a score 5 from our forward model (i.e. not well-aligned). Responses do not undergo rewriting for this baseline.
\item \textit{Dolma + rewriting:} Data obtained from steps (1) and (3) of our pipeline (Figure \ref{fig:main_fig}). We source candidate responses from Dolma, generate instructions with backtranslation and rewrite the responses with Llama-2-70B-chat. No intermediate filtering is done in this case.
\item \textit{Dolma + filtering + rewriting:} Data obtained from going through all the steps of our pipeline (Figure \ref{fig:main_fig}) as described in Section \ref{method}.
\end{itemize}
We find that at this scale, our filtered backtranslation data (\textit{Dolma + filtering}) outperforms similarly constructed data from previous work \cite{li2023self} (\textit{ClueWeb + filtering}). Our best dataset that undergoes both filtering and rewriting yields better win rate than all other baselines. It is worth noting that fine-tuning on rewritten responses from unfiltered instruction-response pairs (\textit{Dolma + rewriting}) outperforms fine-tuning on initial web-scraped responses that have passed the filtering stage but have not been rewritten (\textit{Dolma + filtering}). This signals that the rewriting step is more effective than filtering at improving the quality of instruction-tuning data.
\begin{table}
\centering
\begin{adjustbox}{max width=\textwidth}
\renewcommand{\arraystretch}{1.1}
\small
\hspace{-0.5em}
\begin{tabular}{p{3.55cm}p{1.18cm}p{1.735cm}}
    \hline
    \textbf{Data source} &  \textbf{Data size} & \textbf{Win rate (\%)} \\
    \hline
    Dolma + filtering & 12.8K & 71.70 \\
    Dolma + rewriting & 12.8K & 73.44 \\
    Dolma + filtering + rewriting & 12.8K & \textbf{74.38} \\
    ClueWeb + filtering & 12.8K & 70.77 \\
    Open Orca & 12.8K & 74.20 \\
    ShareGPT & 12.8K & 72.69 \\
    Evol-Instruct & 12.8K & 72.32 \\
    Alpaca-GPT4 & 12.8K & 71.17 \\
    Self-instruct & 12.8K & 65.11 \\
    \hline
\end{tabular}
\end{adjustbox}
\vskip -0.5em
\caption{\textbf{Performance of fine-tuning Llama-2-7B.} Given the same data quantity (12.8K), fine-tuning on the instruction-response pairs obtained from instruction back-and-forth translation outperforms fine-tuning on the backtranslated ClueWeb data from previous work \cite{li2023self}, as well as other common instruction datasets. We also find that the rewriting step is more effective than the filtering step at improving the data quality, and subsequently, the model win rate.}
\label{tab:7b_results}
\vspace{-1em}
\end{table}
\vspace{-0.25em}
\subsection{70B scale}
We also experiment with fine-tuning the Llama-2-70B base model on different variants of the backtranslation data described in the previous section. While the instruction-tuning data generated by previous work \cite{li2023self} is limited by the amount of high-quality text from the initial web corpus (i.e. ClueWeb), our approach overcomes this limitation with response rewriting. We generate 51.2K instruction-response pairs with our pipeline. In Table \ref{tab:70b_results}, we find that by simply doing backtranslation on Dolma texts and filtering like previous work (\textit{Dolma + filtering}), the resulting model slightly lags behind the Humpback model from \cite{li2023self} (\textit{ClueWeb + filtering}) in terms of win rate. However, after rewriting responses in the filtered subset (\textit{Dolma + filtering + rewriting}), we manage to outperform previous work by 3.6\%. Similar to the 7B scale, we also observe at the 70B scale that fine-tuning on rewritten responses from unfiltered instruction-response pairs (\textit{Dolma + rewriting}) is more effective than fine-tuning on web-scraped responses that have passed the filter but have not been rewritten (\textit{Dolma + filtering}).
\begin{table}
\centering
\begin{adjustbox}{max width=\textwidth}
\renewcommand{\arraystretch}{1.1}
\small
\hspace{-0.5em}
\begin{tabular}{p{3.55cm}p{1.18cm}p{1.735cm}}
    \hline
    \textbf{Data source} &  \textbf{Data size} & \textbf{Win rate (\%)}\\
    \hline
    Dolma + filtering & 51.2K & 87.42 \\
    Dolma + rewriting & 51.2K & 90.52 \\
    Dolma + filtering + rewriting & 51.2K & \textbf{91.74} \\
    ClueWeb + filtering & 41.8K & 88.18 \\
    Open Orca & 51.2K & 87.31 \\
    ShareGPT & 51.2K & 88.56 \\
    Evol-Instruct & 51.2K & 86.05 \\
    Alpaca-GPT4 & 51.2K & 86.18 \\
    Self-instruct & 51.2K & 78.48 \\
    \rowcolor{gray!10} Dolma + filtering + rewriting & 25.6K & 90.22 \\
    \rowcolor{gray!10} Dolma + filtering + distilling & 25.6K & 87.58 \\
    \hline
\end{tabular}
\end{adjustbox}
\vskip -0.5em
\caption{\textbf{Performance of fine-tuning Llama-2-70B.} While backtranslation data from previous work \cite{li2023self} is limited by the number of high-quality web pages in Clueweb, our approach relies on Dolma texts and thus has access to many more candidate responses. Similar to the 7B scale results, (i) rewriting is more effective at improving data quality than filtering, (ii) filtering backtranslated instructions and then rewriting the responses does the best and outperforms previous work. We also observe that using an aligned LLM for response rewriting yields better data, and subsequently, win rate, than using the same model for distillation.}
\vspace{-1em}
\label{tab:70b_results}
\end{table}
\vspace{-0.25em}
\section{Understanding rewritten data quality}\label{understanding}
Given the performance benefits of rewritten data, we analyze how the outputs obtained from rewriting are different from those obtained from distillation. We also analyze the characteristics of the instruction-response pairs resulting from our pipeline compared to other existing datasets.
\vspace{-0.25em}
\subsection{Rewriting versus Distilling}\label{rewrite_vs_distill}
As rewriting involves asking Llama-2-70B-chat to improve the response quality, conditioned on an initial web-crawled response and a backtranslated instruction, a fundamental question arises: \emph{does the rewriting process leverage information in the raw text or does it simply distill knowledge stored in Llama-2-70B-chat?}

Given the same set of backtranslated instructions, we use MAUVE score \cite{pillutla2021mauve} to quantify the distributional differences among three sets of responses: initial web-scraped responses (from Dolma), rewritten responses, and responses distilled from Llama-2-70B-chat. MAUVE was originally designed to measure the gap between machine- and human-generated texts. This metric computes the area under the curve of divergence frontiers in a quantized space, after embedding text samples with a language model (by default, GPT-2). MAUVE score ranges between 0 and 1; the higher it is, the more similar the text distributions are. In the first row of Table \ref{tab:mauve_score}, we sample two disjoint sets of 10K distilled responses and find that they exhibit high MAUVE score (0.960) as expected, since they are from the same distribution. Comparing 10K initial responses sourced from Dolma to 10K responses distilled from Llama-2-70B-chat, we observe that these two sets of texts differ significantly, even though they are supposedly responses to the same (backtranslated) queries (MAUVE score = 0.0338). Rewritten responses exhibit some similarity with distilled outputs but there still exists a significant gap between them (MAUVE score = 0.340). 
This suggests that the rewriting process is sufficiently distinct from distillation. We provide some examples of rewritten and distilled responses in Appendix \ref{app:data_examples}.
\begin{table}[]
\centering
\begin{adjustbox}{max width=\textwidth}
\renewcommand{\arraystretch}{1.1}
\small
\begin{tabular}{p{4cm}p{2.1cm}}
    \hline
    \textbf{Text distributions} & \textbf{MAUVE score} \\
    \hline
    Distilled responses vs. \newline Distilled responses & 0.960 $\pm$ 0.002 \\
    \rowcolor{gray!10} 
    Distilled responses vs. \newline Rewritten responses &
0.340 $\pm$ 0.009 \\
    Distilled responses vs. \newline Initial web responses  & 0.0338 $\pm$ 0.0007 \\
    \hline
    \end{tabular}
\end{adjustbox}
\vskip -0.5em
\caption{\textbf{Rewritten responses interpolate between the initial web-scraped responses and the distillation outputs from the aligned LLM used for rewriting.}
We use MAUVE score \cite{pillutla2021mauve} to measure the distances among the three text distributions: initial responses sourced from Dolma, the rewritten responses and the distilled outputs of Llama-2-70B-chat, all in response to the same instruction set. We find that the rewritten responses appear more similar to the distilled outputs compared to the web texts, though there still exists a substantial gap between the first two distributions.
}
\label{tab:mauve_score}
\vspace{-1em}
\end{table}
We also compare empirical performance of fine-tuning on rewritten data versus distilled data. For the latter, we feed 25.6K instructions randomly sampled from our filtered backtranslated dataset to Llama-2-70B-chat and let the model answer directly. For the former, we use the same set of 25.6K instructions and prompt Llama-2-70B-chat to rewrite the corresponding web-scraped responses. Fine-tuning a Llama-2-70B model on the distilled responses yields lower win rate compared to fine-tuning on the rewritten texts (bottom two rows of Table \ref{tab:70b_results}). This demonstrates that the rewriting process improves the quality of response data in general, beyond just extracting what an LLM already knows, possibly because rewriting incorporates the information diversity found in web-scraped texts.

\vspace{-0.25em}
\subsection{Instruction quality analysis}\label{instruction_quality}
\paragraph{Measuring instruction quality empirically.   }
\begin{figure}[t]
\includegraphics[trim=0 0 0 0.4cm,clip,width=0.95\columnwidth]
{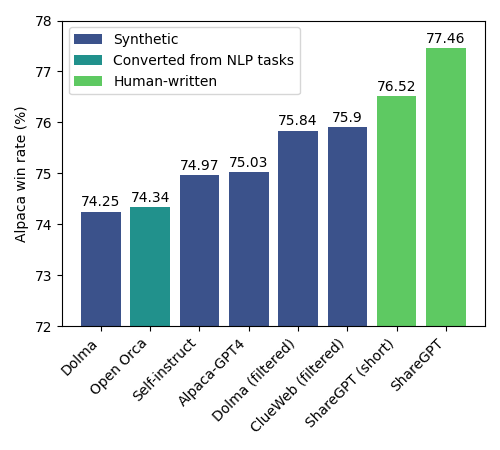}
\vspace{-1.5em}
\caption{\textbf{Quality of instruction prompts from various datasets, measured by their ability to distill useful information from a fixed model.} We randomly sample 12.8K instructions from each dataset in our experiments and input them to Llama-2-70B-chat to distill its knowledge. The quality of the instructions is then measured by the performance of a model (Llama-2-7B) fine-tuned on the (instruction, distilled response) pairs. We find that backtranslated instructions surpass other synthetic instruction generation methods (e.g. Alpaca), while still underperforming human-written queries (e.g. ShareGPT). This gap is partly, but not entirely, due to ShareGPT having longer instructions.}
\label{fig:distill_perf}
\vspace{-1.25em}
\end{figure}
\begin{figure*}[t]
\centering
\includegraphics[trim=0 0 0 0,clip,width=0.92\linewidth]
{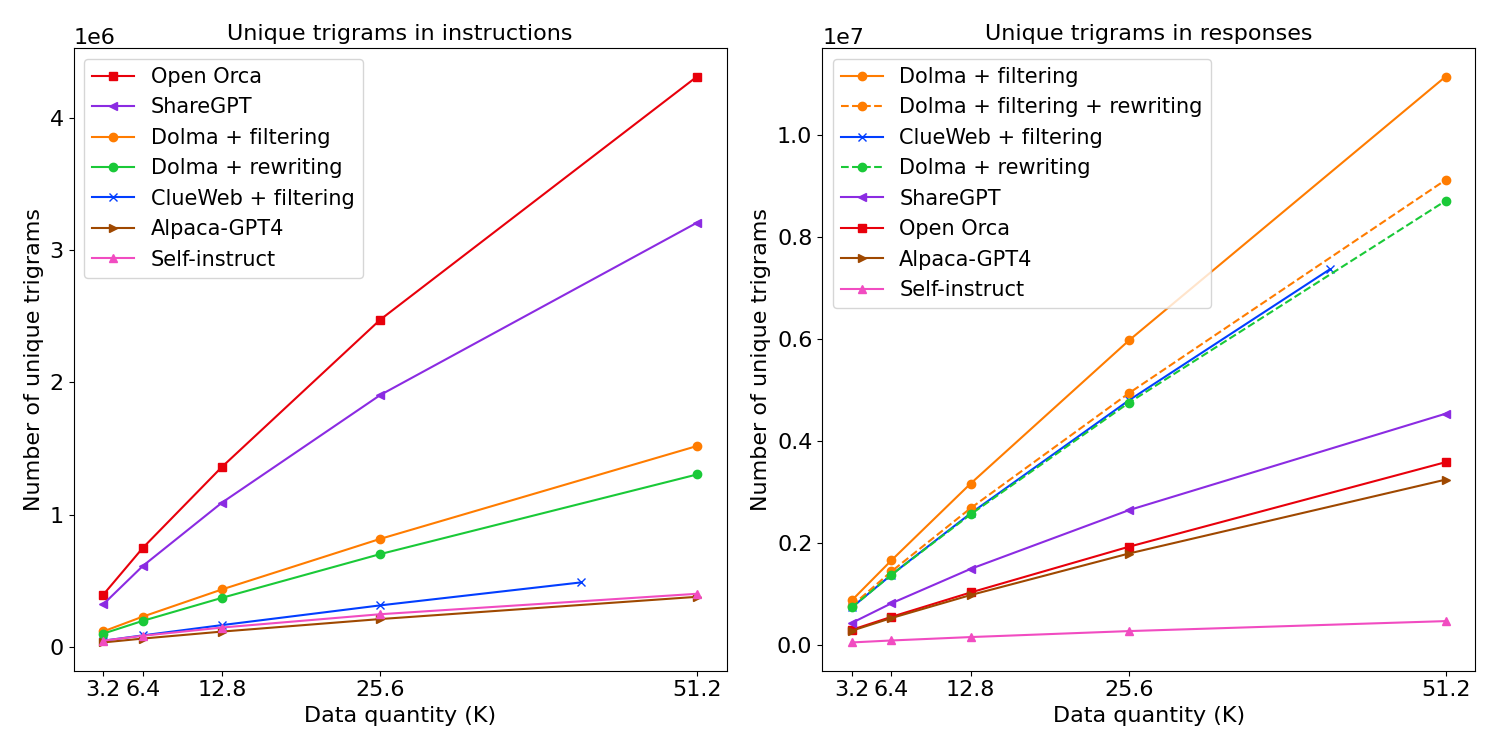}
\vskip -1em
\caption{\textbf{Diversity of instructions and responses from our backtranslation data and other common instruction datasets, as measured by the number of unique trigrams.} We find that \textit{(i)} while instructions generated in this work are more diverse than synthetic instructions from existing baselines including ClueWeb and Alpaca-GPT4, our instruction diversity still lags behind that of manually-crafted prompts, \textit{(ii)} in terms of responses, using web-crawled texts sourced from Dolma yields the most diverse responses; rewriting reduces the response diversity slightly, but still offers much more text diversity than distillation.
}
\label{fig:trigrams}
\vspace{-1em}
\end{figure*}
We attempt to isolate the quality of instructions from different datasets by unifying the response distribution to be outputs distilled from the same model. More specifically, we randomly sample 12.8K instructions from our backtranslation data (with and without filtering) as well as from each of the baseline datasets described in Section \ref{baselines}. We discard all existing answers, and feed each set of 12.8K queries to Llama-2-70B-chat to obtain distilled responses. We then fine-tune a Llama-2-7B model on each set of (instruction, new response) pairs and evaluate the AlpacaEval win rate of the resulting model. In Figure \ref{fig:distill_perf}, we find that \textit{(i)} filtered backtranslation instructions (i.e. from Dolma or ClueWeb) outperform other synthetic instruction generation methods (i.e. Alpaca \cite{taori2023alpaca} and Self-instruct \cite{wang2022self}), \textit{(ii)} however, synthetic instructions still lag behind human-written ones, obtained via user interactions with a chat interface (ChatGPT), \textit{(iii)} instructions constructed from transforming existing NLP datasets (i.e. Open Orca \cite{OpenOrca,mukherjee2023orca}) do not yield high distillation performance, possibly due to limited and repetitive task types.
\vspace{-0.25em}
\paragraph{Diversity.}
As a proxy for diversity, we measure the number of unique trigrams across different instruction sets, given the same data quantity. In Figure \ref{fig:trigrams} (left), we observe that while our instruction set (backtranslated from Dolma) is more diverse than other synthetic instruction sets---including ClueWeb, Alpaca and Self-instruct, there is still a significant gap in diversity between model-generated instructions and manually-crafted ones (e.g., Open Orca and ShareGPT).

It is worth noting that the human-written instructions (e.g. during their interactions with ChatGPT) tend to be substantially longer than synthetic ones, see Table \ref{tab:average_len} in the appendix. We thus conduct an ablation study to study how much the length factor contributes to the quality of instructions. We filter out instructions in ShareGPT that exceed the maximum length of our backtranslated instructions (i.e. 514 tokens) and among the remaining datapoints, randomly sample 12.8K instructions to repeat the distillation experiment described earlier (Figure \ref{fig:distill_perf}). This yields a shorter version of ShareGPT ("ShareGPT (short)") in which the instructions have about the same average length as our backtranslated instructions (i.e. 69 tokens). We observe that when fixing the output distribution to be distilled responses from Llama-2-70B-chat, fine-tuning on "ShareGPT (short)" is less effective than using the original ShareGPT instructions, but still outperforms backtranslated instructions (Figure \ref{fig:distill_perf}). This suggests that independent of length, human-written instructions are still of higher quality overall compared to synthetic instructions.
\vspace{-0.25em}
\paragraph{Complexity.} 
Following the InsTag Complexity metric employed by \citet{liu2023makes}, we use the InsTag public tagger \cite{lu2023instag}, which is a Llama-2-7B model fine-tuned on ChatGPT-generated tags, to automatically tag our text data with all detectable semantics and intentions. We use the average number of unique tags uncovered in instructions, and separately, responses, from each dataset as a proxy for complexity. In Table \ref{tab:average_tags}, we observe that our Dolma-backtranslated instructions offers higher InsTag complexity than most instruction sets from previous work, including ClueWeb-backtranslated data \cite{li2023self} and Open Orca \cite{OpenOrca}. Human-written instructions, i.e. from user interactions with ChatGPT, still yield the highest number of tags on average.
\begin{table}
\centering
\begin{adjustbox}{max width=\textwidth}
\renewcommand{\arraystretch}{1.1}
\small
\begin{tabular}{p{3.6cm}p{1.4cm}p{1.4cm}}
    \hline
    \textbf{Data source} &  \textbf{Instruction} & \textbf{Response} \\
    \hline
    Dolma + filtering & 5.6 & 8.3 \\
    Dolma + rewriting & 4.8 & 6.8 \\
    Dolma + filtering + rewriting & 5.1 & 6.6 \\
    ClueWeb & 3.7 & 6.6 \\
    Open Orca & 3.5 & 4.9 \\
    Alpaca-GPT4 & 3.6 & 4.4 \\
    Self-instruct & 3.1 & 5.1 \\
    ShareGPT & 6.2 & 5.2 \\
    \hline
\end{tabular}
\end{adjustbox}
\vskip -0.75em
\caption{\textbf{Average number of semantic and intention tags uncovered in different instruction-tuning datasets by the InsTag model \cite{lu2023instag}.} While our backtranslated instructions yield fewer tags than human-written ones (i.e. ShareGPT), they exhibit higher complexity than synthetic instructions from other datasets. Besides, we find that responses sourced from or are based on web texts generally have higher InsTag complexity than those obtained via distillation.}
\vspace{-1em}
\label{tab:average_tags}
\end{table}
\vspace{-0.25em}
\subsection{Response quality analysis}\label{response_quality}
\paragraph{Diversity.} We apply the same diversity analysis as in Section \ref{instruction_quality} to response data. In Figure \ref{fig:trigrams} (right), we observe that the initial web responses sourced from Dolma are substantially more diverse than outputs distilled from existing models (e.g. ChatGPT, GPT-4, GPT-3), as well as web texts from ClueWeb. The rewriting process reduces information diversity of these web-scraped responses slightly, but still leads to much more diverse responses than distillation. Overall this analysis demonstrates the importance of including web sources in the data construction process, in order to encourage more diverse instructions \textit{and} responses, compared to just distilling knowledge from existing LLMs. 
\vspace{-0.25em}
\paragraph{Complexity.} 
Applying the same analysis as in Section \ref{instruction_quality}, using the average number of semantic and intention tags uncovered by InsTag tagger \cite{lu2023instag} as a proxy for complexity, we compare the responses from our pipeline (with and without rewriting) to responses from other baseline datasets. In Table \ref{tab:average_tags}, we find that the initial web-scraped responses from Dolma yield the highest complexity. Rewriting generally reduces the InsTag complexity of the response. However, our rewritten responses are still substantially more complex than most existing response data, which is commonly distilled from high-performing LLMs.

We also provide an analysis of instruction and response lengths as another quality metric
. Refer to Appendix \ref{app:data_stats} for more details.
\vspace{-0.25em}
\section{Related Work}
\vspace{-0.25em}
We discuss related papers that construct new instruction-tuning datasets or propose methods to improve existing ones. More in-depth review can be found in \citet{zhang2023instruction}.
\vspace{-0.25em}
\paragraph{Human-crafted data.} Open Assistant \cite{kopf2024openassistant}, Dolly \cite{DatabricksBlog2023DollyV2} and Super-NI \cite{wang2022super} are some examples of datasets that contain solely human-generated and human-annotated conversations, covering a range of topics and NLP tasks. These datasets tend to be relatively small in scale due to the expensive costs of manual annotation and verification.

Other papers do not explicitly ask humans to create questions and answer them, but instead re-purpose existing datasets. For example, FLAN \cite{longpre2023flan} and Natural instructions \cite{mishra2021cross} transform inputs and outputs of more than 60 NLP tasks into instruction-tuning data. This suffers from the same scalability issue as human-annotated datasets.
\vspace{-0.25em}
\paragraph{Synthetic instruction generation.} In contrast to sourcing manually written instructions, which may be expensive to scale, some papers propose ways to automatically generate large quantities of instructions \cite{wang2022self, taori2023alpaca}. In particular, our work is inspired by the backtranslation technique proposed in \citet{li2023self}, which fine-tunes an LLM specifically for the task of instruction generation, and then applies the model to augment text segments extracted from the web with corresponding instructions. The paper suggests that this approach allows the resulting instruction-tuning data to be more diverse especially in the long tail. Another prior work, LongForm \cite{koksal2023longform}, introduces a similar approach for generating instructions.

Most related to our approach is the work by \citet{chen2023tegit}, who train an LLM to generate \textit{both} instructions and responses from web-scraped documents. 
In contrast to their method, we (i) generate instructions separately with backtranslation and then ask an LLM to improve the existing responses, (ii) obtain better performance with much fewer data (Table \ref{tab:cc_sources}) (iii) generate more data (51.2K compared to 12.4K), (iv) offer more insights into the quality of our instructions and responses in comparison to other existing datasets. In addition, concurrent work by \citet{zheng2024kun} also proposes more detailed scoring and refinement prompts to improve the instruction curation and response formatting of the backtranslation pipeline from \citet{li2023self}, applying it to Chinese text data.
\vspace{-0.25em}
\paragraph{Distillation.} Perhaps the most common approach in instruction-tuning data generation, distillation seeks to mimic the capabilities of powerful LLMs (e.g. GPT-4) by feeding queries to these models and using the outputs to fine-tune subsequent LLMs. Datasets that are built this way include ShareGPT \cite{vicuna2023}, OpenInstruct \cite{wang2023far}, Alpaca-GPT4 \cite{peng2023instruction} and UltraFeedback \cite{tunstall2023zephyr}.
\vspace{-0.25em}
\paragraph{Improving instruction-tuning data quality.} Some prior work studies characteristics of high-quality instruction-tuning data \cite{liu2023makes} and proposes curation techniques accordingly. LIMA \cite{zhou2024lima} carefully collects 1K fine-tuning samples via both internet sourcing and human annotation, and shows that strong performance can be achieved despite the small data quantity. Similarly, \citet{chen2023alpagasus} demonstrates that performance gain is possible by fine-tuning on only a small subset of the original dataset (Alpaca), using ChatGPT as the quality evaluator. \citet{zhao2024long} finds that selecting only the 1K longest responses from existing datasets offers a very strong baseline, independent of GPT-4's preference for longer texts. Evol-Instruct \cite{xu2023wizardlm} and Orca \cite{mukherjee2023orca} manually prompt models to enhance the complexity of instructions, and subsequently, data generation (e.g. by asking for justification). \citet{fan2024reformatted} reformats the responses of existing instruction data to augment them with relevant information and align them with pre-determined criteria set by humans.
\vspace{-0.25em}
\section{Discussion}
We propose instruction back-and-forth translation: combining instruction backtranslation method from \citet{li2023self} with response rewriting, in order to benefit from both the information diversity found on the internet and the quality of model annotations, while enabling scalability owing to the size of the web corpus where we source initial responses from. 
\vspace{-1.5em}
\paragraph{Future work.} Our findings motivate a number of interesting future directions. One concrete question is whether applying other existing curation techniques---e.g. quality filters proposed by \citet{liu2023makes}---to our pool of (synthetic instructions, rewritten response) pairs would lead to further performance gains. In addition, we also look forward to scaling up our data generation pipeline and studying the implication of the rewritten data on the pre-training process, given concurrent work \cite{maini2024rephrasing} that explores paraphrasing pre-training data into the question-answering format.
\vspace{-0.25em}
\paragraph{Limitations.} Although we try to control for confounding factors (e.g. data quantity)
, our findings are only obtained from using one model family, i.e. Llama-2 \cite{touvron2023llama2}. Besides, our 
pipeline revolves around general-purpose English instructions, with limited coding or science-related tasks. Nevertheless, it is possible to extend our method to more domain-specific data, e.g. by crawling texts from StackOverflow, generating instructions and rewriting the responses with Code Llama \cite{roziere2023code}. 
\vspace{-0.25em}
\paragraph{Ethical considerations \& Potential risks.} It is possible that sourcing response data from the web could affect the factuality of the fine-tuned model and/ or make it more prone to hallucination. The same risks apply to the response rewriting process. Future work could include additional steps to verify the quality of the information in the responses, and check whether it overlaps with what the model already knows, before using the data for fine-tuning.


\section*{Acknowledgments}
We are grateful to Weijia Shi and Mike Lewis for helpful discussion as well as feedback on the
manuscript. TN is supported by the UW-Meta AI Mentorship Program.
This work is supported in part by NSF grant 2019844.
\bibliography{custom}

\newpage
\appendix
\section{More training details}\label{app:more_train_details}
\subsection{Web corpus ablation}
We randomly sample text segments from different Common-Crawl-derived corpora, generate the corresponding instructions with backtranslation, and filter the candidate pairs to obtain 3200 high-quality samples for fine-tuning Llama-2-7B. In Table \ref{tab:cc_sources}, even though Dolma head split still slightly lags behind ClueWeb when it comes to producing high-quality responses for fine-tuning, it still outperforms other web data sources such as C4 and Dolma middle split.
\begin{table}[h!]
\centering
\begin{adjustbox}{max width=\textwidth}
\renewcommand{\arraystretch}{1.1}
\small
\begin{tabular}{p{3.5cm}p{1.18cm}p{1.74cm}}
    \hline
    \textbf{Data source} &  \textbf{Data size} & \textbf{Win rate (\%)} \\
    \hline
    C4 & 3.2K & 66.83 \\
    Dolma (head) & 3.2K & 67.92 \\ 
    Dolma (middle) & 3.2K & 67.21 \\
    ClueWeb & 3.2K & 68.04 \\
    \hline
\end{tabular}
\end{adjustbox}
\vskip -0.75em
\caption{\textbf{Comparison of text quality from various web-crawled data sources, when used as initial responses in our pipeline.}}
\label{tab:cc_sources}
\end{table}
\subsection{Data generation hyperparameters}\label{app:prompt}
By default, we use nucleus sampling \cite{holtzman2019curious} for our data generation.
\paragraph{(1) Backtranslation} We prompt the fine-tuned backward model to generate instruction that can go with a given web text response, with $T=1.0, p=0.9$. The prompt follows from previous work \cite{li2023self}:
\newline
\newline
\texttt{[INST] Below is a candidate answer to a question or instruction from an user. Write the most likely question to which the text below would be a great answer.
\newline
\newline
<response>
\newline
\newline
Answer in the style of an AI Assistant. [/INST]}
\newline
\paragraph{(2) Filtering} We prompt the fine-tuned forward model to score instruction-response pairs with $T=1.0, p=0.9$. The prompt for scoring could be found in Table 19 of previous work \cite{li2023self}.

\paragraph{(3) Rewriting} Given \texttt{<response>}, which is a cleaned text document from Dolma, and the corresponding \texttt{<instruction>} generated by our backward model, we ask Llama-2-70B-chat to improve the response with $T=1.0, p=0.9$ and the following prompt:
\newline
\newline
\texttt{[INST] Given the draft response to the provided question below, rewrite the draft to improve it, so it is a high quality response to the given question.
\newline
\newline
Draft Response: <response>
\newline
\newline
Question: <instruction> 
\newline
\newline
Given the above question, rewrite the draft response to be an improvement over the draft response. It should be as similar as possible, copying text where possible, while making the flow more clear, useful, relevant and providing a direct answer to the question. It should be written to be impeccably tailored to the user’s question as if written by an AI Assistant, without extraneous information, reflecting expert knowledge, and demonstrating a high-quality, engaging, and insightful answer. Try not to add new facts that are not already in the draft response. Return the rewritten response between [RES] and [/RES]. [/INST]}

\subsection{Training hyperparameters}
We fine-tune Llama-2 model \cite{touvron2023llama2} with 7B and 70B parameters. By default, we use a cosine learning rate schedule with batch size 32, weight decay 0.1 and dropout 0.1. For the 7B scale, we use learning rate 1e-5. With any dataset size smaller than or equal to 25.6K, we fine-tune for \{600, 900, 1200\} steps and report the highest win rate. For the 70B scale, we use learning rate 5.5e-6 and fine-tune on all 51.2K-size datasets for 1600 steps, and all 25.6K-size datasets for 1200 steps.

Following previous work \cite{li2023self}, in each of our fine-tuning experiments, we combine both seed data (3.2K samples from Open Assistant) and the dataset of interest, tagging the former distribution with \texttt{"Answer in the style of an AI Assistant."} and the latter one with \texttt{"Answer with knowledge from web search."} to distinguish the two data sources.

Each 7B-scale fine-tuning run takes 4 hours with 8 A100 GPUs and 12.8K examples, while each 70B-scale run takes 16 hours with 64 A100 GPUs and 51.2K examples.
\subsection{Evaluation hyperparameters}
Similar to \citet{li2023self}, we use nucleus sampling \cite{holtzman2019curious} with temperature $T=0.7, p=0.9$ for generation. When evaluating on the AlpacaEval set \cite{alpaca_eval}, we append the questions with the prompt \texttt{"Answer in the style of an AI Assistant."} or \texttt{"Answer in the style of an AI Assistant, with knowledge from web search if needed."}, in accordance with the tags used during training, and pick whichever prompt that leads to higher win rate on average.

\section{Data statistics}\label{app:data_stats}
\paragraph{Average length}
In Table \ref{tab:average_len}, we report the average length, measured in number of tokens, for instructions and responses from our datasets and other baselines described in Section \ref{baselines}.
\begin{table}[h!]
\centering
\begin{adjustbox}{max width=\textwidth}
\renewcommand{\arraystretch}{1.1}
\small
\begin{tabular}{p{3.6cm}p{1.4cm}p{1.4cm}}
    \hline
    \textbf{Data source} &  \textbf{Instruction length} & \textbf{Response length} \\
    \hline
    Dolma + filtering & 69 & 499 \\
    Dolma + rewriting & 58 & 449 \\
    Dolma + filtering + rewriting & 69 & 468 \\
    Dolma + filtering + distilling & 69 & 567 \\
    ClueWeb & 31 & 442 \\
    Open Orca & 248 & 181 \\
    ShareGPT & 264 & 318 \\
    Alpaca-GPT4 & 25 & 163 \\
    Self-instruct & 38 & 37 \\
    \rowcolor{gray!20} Open Assistant & 40 & 273 \\
    \hline
\end{tabular}
\end{adjustbox}
\vskip -0.75em
\caption{\textbf{Average token length of instructions and responses from different instruction-tuning datasets.}}
\label{tab:average_len}
\end{table}

When it comes to instructions, we find that those generated by backtranslation tend to be longer than those obtained from other synthetic instruction generation methods (i.e. Alpaca \cite{taori2023alpaca} or Self-instruct \cite{wang2022self}). However, our backtranslation instructions are substantially shorter than manually crafted instructions (i.e. Open Orca \cite{mukherjee2023orca,OpenOrca} or ShareGPT \cite{vicuna2023}). We hypothesize that this property of our generated instructions is influenced by the length of instructions in the seed dataset (i.e. Open Assistant) used to train the backward model. Future work could explore using datasets with longer instructions for training the backward model, or enhance the complexity of synthetic instructions further with specifically crafted prompts, such as via the method proposed in Evol-Instruct \cite{xu2023wizardlm}.

In terms of responses, those obtained from instruction back-and-forth translation are overall longer than those found in other existing datasets. We observe that rewriting reduces the response length, possibly due to the removal of irrelevant information in the initial web-crawled responses (468 tokens versus 499 tokens on average). On the other hand, given the same backtranslated instructions, distilling from an existing model yields significantly longer responses than the initial Dolma texts (567 tokens versus 468 tokens on average).

\paragraph{Score distribution}
In step \texttt{(2) Filtering} of our pipeline, we ask the fine-tuned forward model to score (backtranslated instruction, web response from Dolma) pairs. We use a set of regex expressions to extract the scores from the model's responses. We find that the score is valid (i.e. being in the range of 1 to 5) 88\% of the time, with 5-score examples making up 13.6\% of the data pool.
\section{Examples of our instruction-tuning data}\label{app:data_examples}
Refer to Tables \ref{tab:score_5_example}, \ref{tab:score_3_example} and \ref{tab:score_1_example}.
\onecolumn
\begin{small}
\begin{longtable}{p{.25\textwidth} p{.75\textwidth}}
    \hline
    Backtranslated instruction & write a press release for apple announcing the acquisition of shazam \\
    \hline
    Response (from Dolma) & Cupertino, California — Apple today announced it has completed its acquisition of Shazam, one of the world's most popular and highly-rated music apps, used by hundreds of millions of people worldwide. \\
    \hline
    Rewritten response & Apple Acquires Shazam: A New Chapter in Music Recognition Technology
\newline
\newline
Cupertino, California — Apple is thrilled to announce the completion of its acquisition of Shazam, the world-renowned music app used by hundreds of millions of people globally. This strategic move marks a significant milestone in Apple\'s mission to revolutionize the way people discover, enjoy, and interact with music.
\newline
\newline
With Shazam, users can effortlessly identify songs, artists, and albums in a matter of seconds, making it an indispensable tool for music lovers everywhere. By integrating Shazam\'s innovative technology into our ecosystem, we can further enhance the music experience for our customers, providing them with seamless access to the songs and artists they love.
\newline
\newline
"We are delighted to welcome Shazam to the Apple family," said Tim Cook, CEO of Apple. "This acquisition is a testament to our commitment to delivering the best possible music experiences to our customers. We look forward to exploring the exciting possibilities that this partnership presents, and we are eager to see where this new chapter will take us."
\newline
\newline
Shazam has been a game-changer in the music industry since its inception, using audio fingerprinting technology to quickly and accurately recognize songs. This cutting-edge tech has made it possible for users to identify music playing in the background, whether in a store, on the radio, or even in a movie. Shazam has also expanded its offerings to include features like lyrics, recommendations, and playlists, cementing its place as a leader in the music recognition space.
\newline
\newline
The union between Apple and Shazam is a natural one. Both companies share a passion for delivering exceptional user experiences, and we are confident that this collaboration will bring even more joy and convenience to music fans everywhere.
\newline
\newline
"We are incredibly excited to join forces with Apple, a company that shares our dedication to delivering the best possible user experience," said Rich Riley, CEO of Shazam. "We look forward to working together to create new and innovative ways for people to enjoy and engage with music."
\newline
\newline
Apple and Shazam share a rich history, dating back to the early days of the App Store. Shazam was one of the first apps to be featured on the App Store, and it quickly became a fan favorite. Today, this partnership represents the next step in that journey, as we continue to push the boundaries of what is possible with music recognition technology.
\newline
\newline
At Apple, we believe that music has the power to bring people together and enrich their lives. With the addition of Shazam to our family, we are one step closer to making that vision a reality. So, let the music play, and let's see where this journey takes us next.\\
    \hline
    Distilled response & Sure! Here's a possible press release for Apple announcing the acquisition of Shazam:
\newline
\newline
---
\newline
\newline
Apple Acquires Shazam, Bringing Music Recognition Technology In-House
\newline
\newline
Cupertino, California — December 17, 2022 — Apple today announced that it has acquired Shazam, a leading music recognition platform that has been a popular choice for music fans around the world. The acquisition brings Shazam's innovative audio fingerprinting technology and extensive music database into the Apple fold, further enhancing the company\'s capabilities in the music and audio space.
\newline
\newline
"We\'re thrilled to welcome Shazam to the Apple family," said Tim Cook, CEO of Apple. "Shazam's technology and talented team will play a key role in our efforts to continue delivering the best music experiences to our customers. We're looking forward to exploring the many ways that Shazam's audio fingerprinting tech can be integrated into our products and services, and we're excited to see what the future holds for this partnership."
\newline
\newline
Shazam's proprietary audio fingerprinting technology allows users to identify songs and artists playing around them simply by holding their device near the source of the sound. The platform has become a go-to destination for music discovery and identification, with over 1 billion downloads worldwide and counting.
\newline
\newline
"We\'re proud of the Shazam team for building such a powerful and beloved platform," said Oliver Schusser, Vice President of Apple Music. "We're excited to integrate Shazam's technology into our own products and services, and we're looking forward to working with the Shazam team to take music recognition and discovery to the next level."
\newline
\newline
The acquisition follows Apple's purchase of Beats Electronics in 2014, which brought the popular Beats headphones and streaming service into the Apple family. Like that deal, the Shazam acquisition underscores Apple's commitment to delivering the best possible music experiences to its customers.
\newline
\newline
Terms of the deal were not disclosed. Shazam will continue operating independently, with the same leadership and team that have driven its success to date.
\newline
\newline
About Apple
\newline
\newline
Apple revolutionized personal technology with the introduction of the Macintosh computer in 1984. Today, Apple leads the world in innovation with iPhone, iPad, Mac, Apple Watch, and Apple TV. Apple's four software platforms—iOS, macOS, watchOS, and tvOS—provide seamless experiences across all Apple devices and empower people with breakthrough services including the App Store, Apple Music, Apple Pay, and iCloud. Apple's 100,000 employees are dedicated to making the best products on earth, and to leaving the world better than they found it.
\newline
\newline
About Shazam\
\newline
\newline
Shazam is a popular music recognition platform that has been a go-to destination for music fans around the world. With over 1 billion downloads worldwide, Shazam lets users identify songs and artists playing around them simply by holding their device near the source of the sound. Shazam's proprietary audio fingerprinting technology has made it a leader in music recognition and discovery, and its extensive music database has made it a valuable resource for music fans everywhere.
\newline
\newline
Contacts:
\newline
\newline
Apple Media Relations: [media.relations@apple.com](mailto:media.relations@apple.com)
\newline
Shazam Press Office: [press@shazam.com](mailto:press@shazam.com)
\newline
\newline
---
\newline
\newline
Of course, this is just one possible version of a press release announcing Apple's acquisition of Shazam. The actual press release could be different, depending on the specifics of the deal and the messaging that Apple and Shazam want to convey. \\
    \hline
\caption{\textbf{Example of a score-5 (backtranslated instruction, web-scraped response) pair}} \\
\label{tab:score_5_example} \\
\end{longtable}
\end{small}
\twocolumn

\onecolumn
\begin{small}
\begin{longtable}{p{.25\textwidth} p{.75\textwidth}}
    \hline
    Backtranslated instruction & Why doesn't anybody ride the train in San Jose? \\
    \hline
    Response (from Dolma) & Less than 1 percent of Santa Clara County residents ride VTA light rail, and taxpayers subsidize 85 percent of the operating costs.
    \newline
    Light rail recently celebrated its 25th anniversary in Silicon Valley, but most people in the area were probably too busy driving to celebrate. Mike Rosenberg of the Mercury News reports that the system, run by the Valley Transportation Authority, is among the country's worst by just about every common metric of success.
    \newline
    To wit: Less than 1 percent of Santa Clara County residents ride VTA light rail; the per-passenger round-trip operating cost is \$11.74 and taxpayers subsidize 85 percent of costs — third and second worst in the country, respectively. There are problems with measuring costs per passenger mile on light rail, but ouch.
    \newline
    Light-rail agencies in Minneapolis, Houston, Newark, N.J., and Phoenix each run less service than VTA yet carry more passengers than the South Bay's network. Several cities that are much smaller than San Jose — from St. Louis to Salt Lake City to Portland, Ore. — also feature light-rail systems with more riders than VTA.
    \newline
    Sacramento — which also opened its light-rail network in 1987, operates with approximately the same level of service and runs through a similarly sprawled-out region — carries nearly 40 percent more passengers per day than VTA.
    \newline
    When I first encountered VTA light rail in San Jose (briefly) in 1990. I noted how slowly it moved thru the downtown area, intermingling amongst pedestrian traffic. I thought it then to be weird at best for a city to run light rail on a sidewalk where pedestrians normally are. Fast forward to today (after returning to New Jersey then moving back to California), even with system-wide expansion to Mountain View, South San Jose and East San Jose, and its still the same, sad slowness I remember.
    \newline
    Silicon Valley must be getting used to bad transit news at this point. In November, Rosenberg reported that a VTA plan to extend a light rail line 1.6 miles to Los Gatos, home of Netflix, will cost \$175 million while drawing only about 200 new riders. Back in May, a local news station found a culture of fare evasion on VTA that gives the system a rate of 7.2 percent — highest in the region.
    \newline
    In April, construction began on an extension of BART from the Bay Area to Santa Clara, ending at the Berryessa station, where it will meet the VTA light rail system. Among that project's shortcomings are the fact that the stop falls short of downtown San Jose, and that Berryessa seems poorly suited to transit-oriented development. An extension to San Jose is still in the plans, though BART service to the city would duplicate the Caltrain commuter rail that already goes there.
    \newline
    Not that anyone in Silicon Valley uses Caltrain anyway. In October, Stamen Designs released a map of private bus lines that serve the campuses of Google, Apple, Facebook, and the like for employees living in downtown San Francisco. These buses are seen as a better option than Caltrain, which runs infrequently and requires a number of transfers from the city to reach Silicon Valley...
    \\
    \hline
    Rewritten response & San Jose's VTA light rail system has struggled with low ridership, high costs, and poor performance, raising the question of why nobody rides the train in San Jose. According to Mike Rosenberg of the Mercury News, the system has some of the worst metrics in the country, with fewer than 1\% of Santa Clara County residents using the train and taxpayers subsidizing 85\% of the operating costs.
    \newline
    \newline
    One reason for the low ridership may be the limited coverage area, as the system primarily serves the downtown area and nearby neighborhoods. Additionally, the trains move slowly through the downtown area, which may not appeal to commuters looking for a quick and efficient ride.
    \newline
    \newline
    Moreover, Silicon Valley's tech industry leaders seem to prioritize corporate campuses over accessible public transportation. Private bus lines for employees living in downtown San Francisco are viewed as a better alternative to Caltrain, which runs infrequently and requires multiple transfers to reach Silicon Valley.
    \newline
    \newline
    Despite efforts to expand the VTA light rail system, including a planned extension to Los Gatos that will cost \$175 million and draw only 200 new riders, the system still falls short of meeting the region's transportation needs. The Berryessa station, where the VTA light rail system will meet the BART, is not located in downtown San Jose, and the extension to San Jose is still in the plans but will duplicate the Caltrain commuter rail that already exists.
    \newline
    \newline
    In conclusion, the low ridership of the VTA light rail system in San Jose can be attributed to a combination of factors, including limited coverage, slow travel times, and a lack of investment in accessible public transportation by the tech industry. Addressing these issues could potentially improve the system's performance and encourage more people to use the train.\\
    \hline
    Distilled response & There could be several reasons why people may not ride the train in San Jose, California. Here are a few possible reasons:
\newline
\newline
1. Limited public transportation options: San Jose is a sprawling city with a car-centric culture, and public transportation options may be limited, especially outside of downtown areas. This can make it difficult for people to access the train or other forms of public transportation, especially if they live in suburban or rural areas.
\newline
2. Lack of convenient routes: The train routes in San Jose may not be convenient for many residents, especially if they live in areas that are not serviced by the train. This can make it difficult for people to use the train for their daily commutes or other trips.
\newline
3. Limited frequency of service: The frequency of train service in San Jose may be limited, with few trains running during off-peak hours or on weekends. This can make it difficult for people to use the train for spontaneous trips or for traveling outside of peak commute times.
\newline
4. Competition from other modes of transportation: San Jose is located in the heart of Silicon Valley, and many people may prefer to use their own cars or ride-hailing services like Uber or Lyft to get around. This can make it difficult for the train to compete for passengers, especially in areas with heavy traffic or limited public transportation options.
\newline
5. Lack of awareness or promotion: The train service in San Jose may not be well-promoted or well-known, especially among visitors or new residents. This can make it difficult for people to learn about the train and its routes, schedules, and fares.
\newline
6. Affordability: The cost of taking the train in San Jose may be prohibitively expensive for some people, especially for those who are used to cheaper transportation options like driving or ride-hailing services. This can make it difficult for the train to attract a diverse range of passengers.
\newline
7. Safety concerns: Some people may be hesitant to ride the train in San Jose due to safety concerns, such as concerns about crime or safety on the trains or in the surrounding areas.
\newline
8. Lack of amenities: The train service in San Jose may not offer the same level of amenities as other modes of transportation, such as comfortable seating, Wi-Fi, or food and beverage options. \\
    \hline
\caption{\textbf{Example of a score-3 (backtranslated instruction, web-scraped response) pair}} \\
\label{tab:score_3_example} \\
\end{longtable}
\end{small}
\twocolumn

\onecolumn
\begin{small}
\begin{longtable}{p{.25\textwidth} p{.75\textwidth}}
    \hline
    Backtranslated instruction & When was the Chicago Symphony Orchestra's first tour of Russia and Eastern Europe, and were any recordings made during that tour? \\
    \hline
    Response (from Dolma) & You are currently browsing the tag archive for the 'András Schiff' tag.
\newline
Congratulations to Bernard Haitink—the Chicago Symphony Orchestra's principal conductor from 2006 until 2010 and a frequent guest conductor—the recipient of this year's Gramophone magazine award for lifetime achievement!
\newline
Congratulations, Maestro Haitink!
\newline
Haitink returns to Chicago in April 2016, leading the Orchestra in Mozart's Piano Concerto no. 22 with Till Fellner and Strauss's An Alpine Symphony.
\newline
Sir Georg Solti led the Chicago Symphony Orchestra's first trip to the Soviet Union and Hungary in November 1990, also including a single stop in Vienna.
\newline
Two recordings were made during the tour, both for London Records. The performance of Bruckner's Eighth Symphony was recorded live in Leningrad on November 22 (London's first recording venture in the Soviet Union); Michael Haas was the producer, James Lock and Colin Moorfoot were the engineers, and Sally Drew was the tape editor. Mahler's Fifth Symphony was recorded live in Vienna on November 30, Michael Haas was the producer, Stan Goodall was the engineer, and Matthew Hutchinson was the tape editor. \\
    \hline
    Rewritten response & The Chicago Symphony Orchestra's first tour of Russia and Eastern Europe took place in November 1990, led by Sir Georg Solti. During this tour, the orchestra made two recordings for London Records. The first recording was of Bruckner's Eighth Symphony, which was recorded live in Leningrad on November 22, with Michael Haas as producer, James Lock and Colin Moorfoot as engineers, and Sally Drew as tape editor. The second recording was of Mahler's Fifth Symphony, recorded live in Vienna on November 30, with Michael Haas as producer, Stan Goodall as engineer, and Matthew Hutchinson as tape editor.
\newline
\newline
Congratulations are in order for Bernard Haitink, who will be returning to Chicago in April 2016 to lead the orchestra in Mozart's Piano Concerto No. 22 with Till Fellner and Strauss's An Alpine Symphony. Haitink, who served as the principal conductor of the Chicago Symphony Orchestra from 2006 to 2010, is this year's recipient of Gramophone magazine's lifetime achievement award.\\
    \hline
    Distilled response & The Chicago Symphony Orchestra's first tour of Russia and Eastern Europe took place in 1971. The orchestra, led by music director Georg Solti, performed in several cities, including Moscow, Leningrad (now St. Petersburg), Warsaw, and Prague.
\newline
\newline
During the tour, the orchestra made several recordings, including a live recording of Tchaikovsky's Symphony No. 6, which was recorded at the Great Hall of the Moscow Conservatory. This recording was later released on the RCA Red Seal label and won a Grammy Award for Best Classical Performance - Orchestra.
\newline
\newline
Additionally, the orchestra recorded a live performance of Shostakovich's Symphony No. 5 in Leningrad, which was also released on RCA Red Seal.
\newline
\newline
These recordings were significant not only for their artistic quality but also for their historical significance, as they represented one of the first times a Western orchestra had toured the Soviet Union and Eastern Europe since the end of World War II. The tour and recordings helped to establish the Chicago Symphony Orchestra as one of the leading ensembles in the world and paved the way for future cultural exchanges between the United States and Eastern Europe. \\
    \hline
\caption{\textbf{Example of a score-1 (backtranslated instruction, web-scraped response) pair}} \\
\label{tab:score_1_example} \\
\end{longtable}
\end{small}
\twocolumn

\newpage
\section{Rewriting ablations}\label{app:rewrite_ablations}
\begin{figure}[h!]
\centering
\includegraphics[trim=0 0 0 0.4cm,clip,width=\linewidth]
{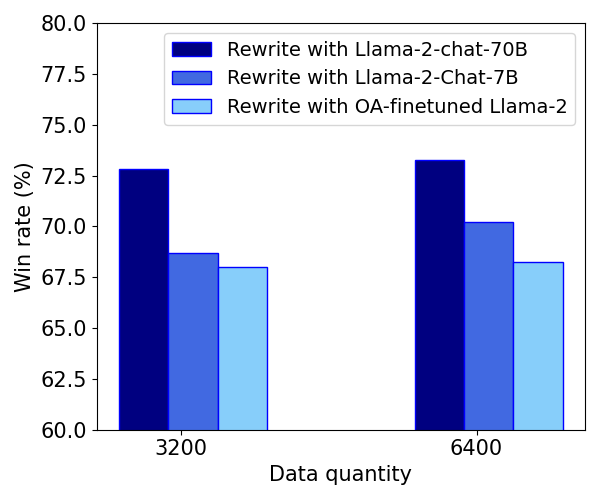}
\vskip -1em
\caption{\textbf{Performance of fine-tuning on responses rewritten by different models.} We find that using a smaller model (i.e. Llama-2-chat-7B) or a less aligned model (i.e. Llama-2 fine-tuned on Open Assistant) for rewriting yields lower response quality, as reflected in the win rate of a Llama-2-7B model fine-tuned on the resulting response data.
}
\label{fig:rewriting_ablations}
\vspace{-1.25em}
\end{figure}

Besides Llama-2-70B-chat, we also experiment with rewriting web-crawled responses with: (i) a smaller  scale model, Llama-2-7B-chat, (ii) a less aligned model, Llama-2 base fine-tuned on Open Assistant seed data (i.e. the forward model used in our filtering step). Ablation results are shown in Figure \ref{fig:rewriting_ablations}.  

In our small-scale experiments fine-tuning a Llama-2-7B model with 3.2K and 6.4K (instruction, rewritten response) pairs, we find that the AlpacaEval win rate is lower when using data rewritten by Llama-2-7B-chat and OA-finetuned Llama-2. This suggests that effectively structuring and enriching the initial raw text response may require a sufficiently aligned model that also contains significant knowledge itself.

\section{AlpacaEval 2.0 results}\label{app:alpaca2}
AlpacaEval 2.0 \cite{dubois2024length} upgrades the baseline model from text-davinci-003 to GPT-4 Turbo, and uses GPT-4 Turbo as the evaluator by default. The evaluation also debiases the raw win rate via a fitted logistic regression model, in order to control for the length of the outputs.

As noted by the Alpaca team, length-controlled (LC) win rates alleviate length biases of GPT-4, but may favor models fine-tuned on its outputs. Our baselines of choice (described in Section \ref{baselines}) mostly distill outputs from GPT4 (in the case of Alpaca-GPT4 and Open Orca) and GPT-related models (ChatGPT for ShareGPT and Evol-Instruct, GPT-3 for Self-instruct). 

We do not emphasize AlpacaEval 2.0 results in the main text, given that the win rates against GPT-4 Turbo obtained from our baselines are close together within a limited range, leading to their performance gaps being not sufficiently significant.

With the current results in Table \ref{tab:70b_results_alpaca2}, we find that similar to the observations in Table \ref{tab:70b_results}: (i) rewriting is more effective at improving data quality than filtering (looking at the first two rows), and (ii) using an aligned LLM for response rewriting yields better win rate than using the same model for distillation (looking at the bottom two rows). However, with this evaluation, rewriting all candidate instruction-response pairs without filtering (\textit{Dolma + rewriting}) yields the best performance, when all baselines are fine-tuned with same data quantity (51.2K).
\begin{table*}[h!]
\centering
\begin{adjustbox}{max width=\textwidth}
\renewcommand{\arraystretch}{1.1}
\small
\hspace{-0.5em}
\begin{tabular}{p{3.55cm}p{1.2cm}p{2.5cm}p{2cm}p{2.5cm}p{2.2cm}}
    \hline
    \textbf{Data source} &  \textbf{Data size} & \textbf{GPT-4-Turbo \newline Win rate - LC (\%)} & \textbf{GPT-4-Turbo Win rate (\%)} & \textbf{Claude-3-Opus Win rate - LC (\%)} & \textbf{Claude-3-Opus Win rate (\%)} \\
    \hline
    Dolma + filtering & 51.2K & 16.71 & 11.90 & 17.66 & 9.63 \\
    Dolma + rewriting & 51.2K & 17.44 & 12.66 & \textbf{21.54} & \textbf{15.34} \\
    Dolma + filtering + rewriting & 51.2K & 16.68 & 12.07 & 19.09 & 12.86 \\
    ClueWeb + filtering & 41.8K & 15.93 & 11.89 & 18.39 & 10.26 \\
    Open Orca & 51.2K & 16.96 & 11.82 & 16.35 & 9.63 \\
    ShareGPT & 51.2K & \textbf{18.42} & 11.62 & 19.14 & 9.81 \\ 
    Evol-Instruct & 51.2K & 15.65 & 9.51 & 17.89 & 8.94 \\ 
    Alpaca-GPT4 & 51.2K & 17.19 & 11.48 & 17.04 & 8.63 \\
    Self-instruct & 51.2K & 12.20 & 7.09 & 13.24 & 6.58 \\
    \rowcolor{gray!10} Dolma + filtering + rewriting & 25.6K & 17.90 & \textbf{13.05} & 17.55 & 11.86 \\
    \rowcolor{gray!10} Dolma + filtering + distilling & 25.6K & 17.80 & 11.95 & 17.50 & 10.50 \\
    \hline
\end{tabular}
\end{adjustbox}
\vskip -0.5em
\caption{\textbf{Win rates against GPT-4-Turbo of Llama-2-70B models fine-tuned on different instruction datasets, as evaluated with AlpacaEval 2.0 framework using GPT-4 Turbo and Claude-3 Opus as evaluators.}}
\label{tab:70b_results_alpaca2}
\end{table*}

\section{Other NLP evaluations}\label{app:other_nlp_evals}
\begin{table*}[h!]
\centering
\begin{adjustbox}{max width=\textwidth}
\renewcommand{\arraystretch}{1.1}
\small
\begin{tabular}{p{4.0cm}p{1.4cm}p{1.6cm}p{1.2cm}p{1.2cm}p{1.2cm}}
    \hline
    \textbf{Data source} &  \textbf{Data size} & \textbf{HellaSwag} & \textbf{ARC} & \textbf{PIQA} & \textbf{MMLU} \\
    \rowcolor{gray!20} \multicolumn{6}{c}{7B scale} \\
    Dolma + filtering & 12.8K & \textbf{60.6} & 59.1 & \textbf{76.1} & 35.6 \\
    Dolma + rewriting & 12.8K & 56.4 & 56.8 & 74.6 & 33.8 \\
    Dolma + filtering + rewriting & 12.8K & 56.8 & 56.7 & 74.5 & 36.6 \\
    ClueWeb + filtering & 12.8K & 60.3 & 57.5 & 75.7 & 35.1 \\
    Open Orca & 12.8K & 57.3 & 59.1 & 76.0 & \textbf{45.7} \\
    ShareGPT & 12.8K & 56.3 & \textbf{59.2} & 75.1 & 40.4 \\
    Evol-Instruct & 12.8K & 54.2 & 56.1 & 75.3 & 35.9 \\
    Alpaca-GPT4 & 12.8K & 57.3 & 57.7 & 74.7 & 35.0 \\
    Self-instruct & 12.8K & 56.7 & 55.7 & 74.9 & 35.2 \\
    \rowcolor{gray!20} \multicolumn{6}{c}{70B scale} \\
    Dolma + filtering & 51.2K & \textbf{68.1} & 69.4 & 81.8 & 62.8 \\
    Dolma + rewriting & 51.2K & 66.4 & 69.3 & 81.2 & 61.9 \\
    Dolma + filtering + rewriting & 51.2K & 66.8 & 69.1 & 81.4 & 61.4 \\
    ClueWeb + filtering & 41.8K & 68.8 & 68.4 & 81.6 & \textbf{63.3} \\
    Open Orca & 51.2K & 65.8 & 66.6 & 78.6 & 61.8 \\
    ShareGPT & 51.2K & 65.7 & 68.4 & 79.7 & 60.0 \\
    Evol-Instruct & 51.2K & 66.8 & 68.5 & 80.4 & 60.9 \\
    Dolma + filtering + rewriting & 25.6K & 66.8 & 69.7 & \textbf{82.2} & 62.7 \\
    Dolma + filtering + distilling & 25.6K & 65.7 & \textbf{70.9} & 82.0 & 60.8 \\
    \hline
\end{tabular}
\end{adjustbox}
\vskip -0.5em
\caption{\textbf{Performance of our fine-tuned models on different NLP tasks.}}
\label{tab:nlp_results}
\end{table*}
We also evaluate the models described in Section \ref{performance} on some common NLP benchmarks:
\begin{itemize}[topsep=0pt, itemsep=0pt, leftmargin=8pt, parsep=2pt]
\item HellaSwag \cite{zellers2019hellaswag}: consists of 70K multiple-choice questions designed to test grounded commonsense inference. Each question comes from either activitynet or wikihow, along with four answer choices about what might happen next in the scene. The correct answer is the actual sentence for the next event, while the other three are adversarially generated and human verified.
\item ARC \cite{clark2018think}: aims to test advanced question-answering capabilities with 7787 grade-school level, multiple-choice science questions. The dataset consists of Challenge and Easy Sets, with the former containing only questions answered incorrectly by both a retrieval-based algorithm and a word co-occurrence algorithm.
\item PIQA \cite{bisk2020piqa}: another multiple-choice dataset, created to test an NLP model's understanding of the physics model of the world. Questions are inspired by how-to instructions and the model is supposed to pick the correct answer out of two choices.

\item MMLU \cite{hendrycks2020measuring}: 
covers 57 subjects across STEM, the humanities, the social sciences, etc. The question difficulty ranges from an elementary level to an advanced professional level, testing for both world knowledge and problem solving ability. Questions are presented in a multiple-choice format.
\end{itemize}
We report the results on these tasks in Table \ref{tab:7b_results}.
\end{document}